\documentclass{article}
\usepackage{graphicx}
\usepackage{etoolbox}
\AtBeginEnvironment{quote}{\par\small}

\usepackage[backend=biber,natbib=true,style=numeric]{biblatex}

\bibliography{annalspaper.bib}
\title{ELIZA Reinterpreted: The world's first chatbot was not intended as a chatbot at all}
\usepackage{authblk}
\author[1]{Jeff Shrager}
\affil[1]{Blue Dot Change and Stanford University Symbolic Systems Program (Adjunct); jshrager@stanford.edu}
\date{April 2024}

\begin{document}

\maketitle

\begin{abstract}
ELIZA, often considered the world's first chatbot, was written by Joseph Weizenbaum in the early 1960s.  Weizenbaum did not intend to invent the chatbot, but rather to build a platform for research into human-machine conversation and the important cognitive processes of interpretation and misinterpretation. His purpose was obscured by ELIZA's fame resulting, in large part, from the fortuitous timing of it's creation and it's escape into the wild. In this paper I provide a rich historical context for ELIZA's creation, demonstrating that ELIZA arose from the intersection of some of the central threads in the technical history of AI. I also briefly discuss how ELIZA escaped into the world, and how its accidental escape, along with several coincidental turns of the programming language screws, led both to the misapprehension that ELIZA was intended as a chatbot, and to the loss of the original ELIZA to history for over 50 years.
\end{abstract}

``We can only see a short distance ahead, but we can see plenty there that needs to be done.'' (The last line of Turing's 1950 MIND paper \cite{AT50Mind})

\section{Introduction}

ELIZA, often considered the world's first chatbot, was written by Joseph Weizenbaum at MIT in the early 1960s.\cite{JW66ELIZA} In building ELIZA, Weizenbaum did not intend to invent the chatbot.\footnote{Of course, it would not have been called a ``chatbot'' at the that time anyway, as that term was not invented until the mid 1990's\cite{chatbot}, but we'll use that term here as it is the currently appropriate term.} Instead, he intended to build a platform for research into human-machine conversation. This may seem obvious -- after all, the title of Weizenbaum's 1966 CACM paper is ``ELIZA-- A Computer Program For the Study of Natural Language Communication Between Man And Machine.'', not, for example, ``ELIZA - A Computer Program that Engages in Conversation with a Human User''. But Weizenbaum's purpose for ELIZA was obscured by the circumstances of its creation, and by its own fame resulting, in large part, from the fortuitous timing of it's creation and it's escape into the wild. 

In this paper I try to provide a rich historical context for ELIZA's creation. ELIZA arose from the intersection of some of the central threads in the technical history of AI. In addition to explaining how these intersect in Weizenbaum's creation of ELIZA, I also briefly discuss how ELIZA escaped into the world through no action or intention of Weizenbaum's, and how its accidental escape, along with a coincidental turn of the programming language screw, led to the misapprehension that ELIZA was intended as a chatbot.  

\section{Why ELIZA?}

Why Joseph Weizenbaum built ELIZA does not appear to be much of a mystery; Isn't the answer just as simple as Pamela McCorduck put it in Machines Who Think\cite{PM79MWT}:

\begin{quote}
``[...] Weizenbaum got interested in language. Ed Feigenbaum introduced him to [...] Kenneth Colby, a psychiatrist who had [...] turned to computers as a possible way of gaining new insights into neurotic behavior. [...] In 1963, Weizenbaum went to MIT, [where] he designed a program that would answer [simple questions]. It was a short, tricky program, based on sleight of hand, and it led Weizenbaum to ask himself some very serious questions about mystification and the computer [...]. [If] you could do a simple question-answering machine, why not a complicated one? How different would complexity make such a machine? Could you seem to have complex responses based on simple rules? [...]  Weizenbaum drove into work many a morning with his neighbor Victor Yngve, who had developed the COMIT language, for pattern matching. If you were going to play around with matching patterns, why not the patterns in English words and sentences? ELIZA was the result. ELIZA was intended to simulate—or caricature, as Weizenbaum himself suggests—the conversation between a Rogerian psychoanalyst and a patient, with the machine in the role of analyst.'' 
\end{quote}

The reason that Weizenbaum gives McCorduck for choosing the Rogerian setting was the simplicity of creating an ``illusion of mutual understanding'': 

\begin{quote}
``What I mean here is the cocktail party conversation. Someone says something to you that you really don't fully understand, but because of the context and lots of other things, you are in fact able to give a response which appears appropriate, and in fact the conversation continues for quite a long time. We do it all the time, not only at cocktail parties. Indeed, I think it's a very necessary mechanism, because we can't, even in serious discussion, probe to the limit of possible understanding. [...] That's necessary. It's not cheating.''\cite[251-253]{PM79MWT}
\end{quote}

When Weizenbaum was looking for a context where he could carry on that sort of illusion, he needed one where ignorance would not destroy the illusion of understanding: ``For example [he goes on], in the psychiatric interview the psychiatrist says, tell me about the fishing fleet in San Francisco. One doesn't say, ``Look, he's a smart man—how come he doesn't know about the fishing fleet in San Francisco?'' What he really wants to hear is what the patient has to say about it. [...]''\cite[251-253]{PM79MWT} [italics as in the original]

However, as mentioned above, the title of Weizenbaum's 1966 paper belies -- or at least complexifies -- the post hoc account he gave McCorduck. The title of the paper is not ``ELIZA -- A Computer Program that Simulates or Caricatures the conversation between a Rogerian psychoanalyst and a patient.'' but rather: ``ELIZA -- A Computer Program For the Study of Natural Language Communication Between Man And Machine''.\cite{JW66ELIZA} This, and other features of that paper that I go into below, as well as a recently discovered manuscript signed in Weizenbaum's own hand\cite{JWELIZAFutureDraft}, suggest that ELIZA was not, as quoted by McCorduck ``intended to simulate—or caricature [...] the conversation between a Rogerian psychoanalyst and a patient'', but rather that it was intended, just as the title of the paper says, as a platform for research in [human]-machine communication. To understand the details of this misinterpretation, we need to understand the context in which ELIZA appeared, and how ELIZA came it into the public eye. 

\section{The Intelligence Engineers}

The founding father of the effort to build intelligent machines was, of course, Alan Turing (although Ada Lovelace may have been the founding mother, as will be seen shortly). Turing is publicly most famous for having led the team that built the ``bombe'', the machine that made it possible to decipher enemy messages in the second World War.\cite{Dyson2012} However, in academic circles, Turing is best known for the mathematical construct that now bears his name, the Universal Turing Machine. And in AI circles  -- and recently in public, as AI has come more into public focus -- Turning is associated with the Imitation Game, which we now call the Turing Test.\cite{AT50Mind}

Since we are discussing ELIZA, which was apparently a chatbot, the reader might expect me to go straight to the Turing Test. However, I am concerned here primarily with ELIZA as a computational artifact, not with whether ELIZA was or was not intelligent; We can all agree that it was not, and we shall see later that Weizenbaum himself had specific reasons for \textit{not} thinking of ELIZA as intelligent. Thus the Turing Test plays little role with respect to ELIZA, at least as it that program was conceived by Weizenbaum, so we shall leave it aside in the present exploration, except for one important detail that arises, almost in passing, in Turing's Mind paper. 

Turing notes that over a century before his effort, Ada Lovelace had described the potential of Babbage's Analytical Engine to ``act upon other things besides number[s]'': 

\begin{quote}
    ``The operating mechanism [...] might act upon other things besides number, were objects found whose mutual fundamental relations could be expressed by those of the abstract science of operations, and which should be also susceptible of adaptations to the action of the operating notation and mechanism of the engine. Supposing, for instance, that the fundamental relations of pitched sounds in the science of harmony and of musical composition were susceptible of such expression and adaptations, the engine might compose elaborate and scientific pieces of music of any degree of complexity or extent [...]''\cite{ADA43}
\end{quote}

We shall soon see that Ada's insight that machines could act upon other things besides numbers was enormously prescient, foreshadowing the concept of symbolic computing, which was to become, and remains today, one of the foundations of artificial intelligence, and which is central to the pre-history of ELIZA. 

But first let us return to Turing's mathematical contribution, Turing's Universal Machine, the now so-called ``Turing Machine''. The Turing Machine was Turing's contribution to a line of inquiry in the early part of the 20th century which included Turning, Kurt Godel, and Alonso Church, all addressing a problem posed in 1928 by Hilbert and Ackerman called the ``Entscheidungsproblem''.\cite{sephilbert} Hilbert's challenge was to find an algorithm to determine whether a mathematical proposition is provable. Godel addressed this problem through what we now call ``Godel Numbering'', his method of turning mathematical expressions into numbers, and showing that there are expressions that cannot be proved in a complete system. Church reached the same conclusion by formalizing general recursive functions in a system called the ``Lambda Calculus''. And Turing addressed the problem by describing a universal machine that can compute any function. Turing then showed, equivalently to Godel and Church, that there are programs for which it is impossible to prove that they will come to a halt on his machine, which we now call ``the halting problem''.\cite{CP08AT}

The connection between Turing's interests in universal computers, computable functions, and intelligence is obvious: If a machine can compute any function (leaving aside provability, which is not prima facie relevant as regards human intelligence), and intelligence is supposed to be some sort of function (which may be separately debated, but which was assumed by Turing), then if one's goal is to understand intelligence and perhaps even to create an intelligent machine, it is important to be able to tell when you've succeeded -- thus, Turing's invention of The Imitation Game, now famously called the ``Turing Test''. 

Turing's Universal Machine was a theoretical construct. But in 1943, he designed and succeeded in building, with other engineers at Bletchley Park, a near-instantiation of his theoretical machine, called Colossus. Dyson observes that Colossus ``was an electronic Turing machine, and if not yet universal, it had all the elements in place.''\cite[256]{Dyson2012} The Turing Machine, and its instantiation in Colossus, represented an extremely important advance in computing, even though computing was only in its infancy. Up until that time, all computers were ``hardwired'' -- that is, they executed a program that was wired into their hardware. The bombe was of this type. The important conceptual advance due to Turing, and at about the same time John von Neumann in the EDVAC project, was to store the program on a medium that is not fixed, but can be changed by the program itself. In Turning's machine (and Colossus) this was a tape. In von Neumann's case it was an electronic memory\cite{WPVNA}. This way of thinking about computers -- as ``stored program'', as opposed to ``hardwired'' -- was the  engineering revolution hidden in Turing's theoretical construct -- that computers could not only do complex calculations, but that they could \textit{manipulate their own programs, much like intelligent agents engage in reasoning, planning, and other meta-cognitive activities wherein we think about and modify our own thoughts}. We now call such machines ``von Neumann-style'' machines, although they could just as reasonably, and perhaps even more so, be called ``Turing-style''. 

\section{Newell, Shaw, and Simon's IPL Logic Theorist: The First True AIs}

Beginning in 1950, almost exactly at the same time that Turing's Mind paper appeared, there was an explosion in von Neumann/Turing style computer hardware. At that time, all of the existing computers, such as Colossus and EDVAC, were one-off machines. Although UNIVAC, the company spun off from Penn's ENIAC project by Eckert and Mauchley, had promised a mass-produced machine, none had yet been delivered when RAND decided to build the JOHNNIAC.\cite{RANDHistory} This was the machine on which Shaw, Newell, and Simon pioneered the IPL series of programming languages, which they used to implement what were are almost certainly the first real AIs: 

\begin{quote}
    ``IPL was first utilized to demonstrate that the theorems in Principia Mathematica which were proven laboriously by hand, by Bertrand Russell and Alfred North Whitehead, could in fact be proven by computation. According to Simon's autobiography Models of My Life, this application was originally developed first by hand simulation, using his children as the computing elements, while writing on and holding up note cards as the registers which contained the state variables of the program. IPL was used to implement several early artificial intelligence programs, also by the same authors: the Logic Theorist (1956), the General Problem Solver (1957), and their computer chess program NSS (1958). Several versions of IPL were created: IPL-I (never implemented), IPL-II (1957 for JOHNNIAC), IPL-III (existed briefly), IPL-IV, IPL-V (1958, for IBM 650, IBM 704, IBM 7090, Philco model 212, many others.''\cite{WPIPL}
\end{quote}

IPL introduced what became the most important concepts in classical Artificial Intelligence (and, in fact, in computing more generally), including list processing, symbolic computing (as foreshadowed by Ada Lovelace!), and recursion.\cite{WPIPL}  

Notably, although Turing's bombe, the code-breaking machine, was not a ``modern'' computer, in the sense that is was hardwired, as opposed to being of the von Neumann/Turing stored program type, the program that it ran was explicitly engaged in mechanical heuristic search, that is, in trying a huge number of operators (e.g., combinations of code keys) in a cleverly arranged order (e.g., beginning with the most probable), in hopes of reaching a goal (e.g., the key that would break the day's code). Much later, Newell and Simon described heuristic search as a fundamental process of intelligence,\cite{NEWELLSIMON81}, and remains one of the most important tools of AI.\footnote{Allen Newell once told me, in passing, that AI's only had one tool, the ability to search huge multi-dimensional spaces efficiently, but that one tool is all you need.} 

However, IPL was an ugly programming language -- not very far from assembly language -- and, as an interpreted language it was slower than assembly or the soon-to-follow ``high level'' compiled languages such as COBOL and Fortran. These promised programmers both high performance and simplicity of expression. This brings us to SLIP and Lisp, which took different paths to bring simplicity and additional power to the important AI concepts pioneered in IPL.

\section{From IPL to SLIP and Lisp} 

Recall (from McCorduck) that Weizenbaum was connected to AI through a number of paths, including Kenneth Colby, a Stanford psychiatrist who was interested in modeling neurosis and paranoia\cite{C&G64}, and Ed Feigenbaum, a computer scientist at Berkeley. Feigenbaum had been a student of Simon's, and was creating various AI programs in IPL.\cite{Feigenbaum1959} Once Weizenbaum got to MIT, he became associated with Project MAC, which was begun, in part, by John McCarthy, the inventor of Lisp and the person who coined the term ``Artificial Intelligence''\cite{WPDartAIConf}.\footnote{Although McCarthy moved to Stanford in 1962, two years before Weizenbaum arrived at MIT.\cite{WPJM}} 

But Weizenbaum was, first and foremost, what we would now call a software engineer\footnote{That term was coined in the late 1960s.\cite{Cameron2018}}, having just come from GE where he worked on highly practical programs. There were several projects mounted to build on the clear successes of Newell and Simon's IPL work without having to cope with its ugliness and inefficiency. As mentioned above, there were already several much more programmer-friendly languages, notably Fortran and COBOL. But these languages were aimed at science, engineering, and business, and did not provide the AI-related functionalities of IPL, such as symbol processing, lists, and recursion. So the question naturally arose as to how to add these capabilities to those already-existing languages. 

The inventive entanglement between Fortran, IPL, and Lisp is concisely captured in a brief mention by Gelernter and coworkers in creating FLPL, the ``Fortran List Processing Language'':

\begin{quote}
``[...] consideration was given to the translation of a JOHNNIAC IPL for use with the IBM 704 computer. However, J. McCarthy, who was then consulting for the project, suggested that Fortran could be adapted to serve the same purpose. He pointed out that the nesting of functions that is allowed within the Fortran format makes possible the construction of elaborate information-processing subroutines with a single statement. The authors have since discovered [...] the close analogy that exists between the structure of [a Newell, Shaw, and Simon] list and a certain class of algebraic expressions that may be written within the language. [...] Not to be overlooked is the considerable sophistication incorporated into the Fortran compiler itself, all of which carries over, of course, into our Fortran-compiled list-processing language. It is reasonable to estimate that a routine written in our language would run about five times as fast as the same program written in an interpretive language [like IPL].''\cite[88-89]{HG60FLPL}  
\end{quote}

Like Gelernter, Weizenbaum implemented a set of IPL-inspired list-processing facilities as a set of Fortran-(and later MAD-)-callable functions, which he called SLIP. In creating SLIP, Weizenbaum took a route similar to Gelernter's. In his 1963 paper, Weizenbaum (in addition to citing IPL and FLPL as influences) critiques McCarthy's Lisp (although not mentioning it by name):

\begin{quote}
``List processing has won a number of dedicated converts. Some have, however, become somewhat too fervent in their advocacy of list processing. While there may be some programming tasks which are best solved entirely within some list processing system, most tasks coming to the ordinary programmer require the application of a number of distinct techniques. The packaging of a variety of tools within a single tool box appears to be a good, if not an optimum, way of outfitting a worker setting out to solve complex problems. FORTRAN, ALGOL, and other languages of the same type provide excellent vehicles for such provisioning. Apart from the fact that they are very powerful in themselves, they have the advantage that they are well known. The task of coming to grips with these new techniques is then that of adding to a vocabulary of an already assimilated language rather than that of learning an entirely new one.''\cite[535-536]{JW63SLIP}
\end{quote}

The acknowledgement in the 1963 SLIP paper is worth quoting in full, as it makes many of the connections explicit: 

\begin{quote}
``[...] SLIP owes a considerable debt to previous list processing systems. Certain of its features are, however, more the result of attempts to build a symbol manipulator for the use of behavioral scientists than to generalize other processors. In this connection, the continuing and generous advice and support of Kenneth Colby, M.D., of Stanford University and of Dr. Edward Feigenbaum of the University of California at Berkeley is gratefully acknowledged. The author also wishes to thank Howard Sturgis of the University of California at Berkeley and Larry Breed of Stanford University for their parts in making the system operative on the computers at their respective computation centers.''\cite[536]{JW63SLIP}\footnote{Symbols in Lisp and IPL-V are named pointers and those languages, being self-contained, have specific support for symbol manipulation. Because SLIP is embedded in another language (initially Fortran, later MAD in which ELIZA was implemented), the naming, should the programmer choose to do so, results from Fortran assignment to variables. It would require another whole paper to track the meaning of ``symbol'' through the history of programming languages, AI, and cognitive science.
}
\end{quote}

Although the SLIP paper makes no mention of ELIZA, nor of any specific application, this acknowledgement makes clear that SLIP was motivated by ``previous list processing systems'' (specifically IPL-V, and FLPL), as a ``symbol manipulator for the use of behavioral scientists'', and that SLIP was running on computers at both Stanford and Berkeley. Although the SLIP paper explicitly cites Gelernter's FLPL, it is likely, given the above acknowledgement, close timing, and length of publication cycles, that SLIP was developed essentially simultaneously with FLPL, and that the citation was a publication-sequence nod, rather than FLPL having directly influenced SLIP's development. Whereas Weizenbaum originally embedded SLIP for Fortran, the SLIP that ELIZA is written in was embedded in MAD (although using the same underlying foreign-function calling machinery as Fortran -- an intermediate language called ``FAP'' -- the Fortran Assembly Program). Intriguingly, this SLIP implementation for the 7090 MAD programming language came from Yale\cite[62L1]{MADSLIPManual}, suggesting some interaction with Gelernter, although the details are obscure. 

Notable from the above is McCarthy's involvement with FLPL, and especially the point made by McCarthy about functional nesting of Fortran statements. McCarthy's own approach to creating a high-level AI language was very different from the one he recommended to Gelernter, which was also adopted by Weizenbaum. In addition to his interest in formal matters of logic and mathematics, McCarthy was coincidentally playing around with Church's lambda calculus, which, as discussed above, was the motivation for Turing's work that led to universal computers. This aligned nicely with the functional nesting pointed out by McCarthy to Gelernter. Recursion is central to both universal computation and AI.\cite{GEB} McCarthy, along with his students, most notably, Steve Russell, figured out, perhaps coincidentally, how to bring list processing and recursion together into an elegant programming paradigm that was far simpler and more elegant than IPL, FLPL or SLIP; thus, the birth of Lisp, among the world's most influential and enduring programming languages.\cite{ChisnallLisp2011}

\section{A Critical Tangent into Gomoku}

Before getting to ELIZA itself, It will be useful to take a brief look at an obscure 1962 paper of Weizenbaum's -- his first -- published in the trade magazine, Datamation.\cite{JW1962Gomoku} Still with GE at the time, this brief paper has the odd and telling name: ``How to make a computer \textit{appear} intelligent'' [emphasis as in the original]. The article, only a bit over two pages long, reports a simple strategy for playing gomoku -- a GO-like game. Weizenbaum didn't actually write the program described in the paper, although he apparently designed the algorithm\footnote{``A program implementing the strategy here outlined has been written by R. C. Shepardson''\cite[26]{JW1962Gomoku}} Regardless, Weizenbaum's interest in this program is in not the algorithm itself, which he describes as ``simple''. Rather, he is interested in the fact that the program was able to ``create and maintain a wonderful illusion of spontaneity'' (p. 24). Indeed, the paper (again, only a bit over two pages long) spends the first half page presenting a caustic screed against AI that is worth reproducing in full: 

\begin{quote}
``There exists a continuum of opinions on what constitutes intelligence, hence on what constitutes artificial intelligence. Perhaps most workers in the fields of heuristic programming, artificial intelligence, et al, now agree that the pursuit of a definition in this area is, at least for the time being, a sterile activity. No operationally significant contributions can be expected from the abstract contemplation of this particular semantic navel. Minsky has suggested in a number of talks that an activity which produces results in a way which does not appear understandable to a particular observer will appear to that observer to be somehow intelligent, or at least intelligently motivated. When that observer finally begins to understand what has been going on, he often has a feeling of having been fooled a little. He then pronounces the heretofore ``intelligent'' behavior he has been observing as being ``merely mechanical'' or ``algorithmic.'' The author of an ``artificially intelligent'' program is, by the above reasoning, clearly setting out to fool some observers for some time. His success can be measured by the percentage of the exposed observers who have been fooled multiplied by the length of time they have failed to catch on. Programs which become so complex (either by themselves, e.g. learning programs, or by virtue of the author's poor documentation and debugging habits) that the author himself loses track, obviously have the highest IQ's.''\cite[24]{JW1962Gomoku}
\end{quote}

The paper then goes on to describe the ``simple'' algorithm in typical algorithmic terms, and at the end, far from closing the loop on his opening salvo, simply suggests some ways to potentially improve the program's play. 

This curious paper provides a deep and interesting insight into the dual forces tearing at Weizenbaum and leading to ELIZA. He is, first and foremost, a software engineer. However, here we have a paper about a game-playing program that begins with a screed about how AI engineers are out to fool people. Weizenbaum clearly has a direct interest in AI, and he does not think much of it. Specifically, he is concerned about how easily people can be ``fooled'' by complex programs into the ``illusion'' of intelligence. A more generous and interesting way to put this is that \textit{Weizenbaum is focused on the users, not on the programs}. Specifically, what is interesting to him is not AI per se, which Weizenbaum has only a little to say about, and not much good, but the human psychological process of interpretation -- in this case, how the users interpret a ``simple algorithm'' as playing intelligently. 

\section{Interpretation is the Core of Intelligence}

Interpretation is the process through which humans -- one might say, intelligences of whatever sort, or, more generally, ``cognitive agents'' -- assign meaning to experiences. By ``meaning'' here we mean approximately what linguists mean by the term ``semantics'' or cognitive scientists mean by ``mental models''. It is the ability to assign meaning, separate from the experience itself, and thence to draw inferences based upon this assignment, that enables cognitive agents to, at the same time, abstract away from the specifics of experiences, and reason about those specifics.\footnote{Of course, terms such as ``meaning'' and ``experience'' are vague, and in a paper focused on interpretation these would need to be specified more carefully, but they are not central to the present exploration. For a broad review, see \cite{SEoPmeaning}.}  

The importance of interpretation has been rediscovered in both psychology and AI dozens of times under almost as many names, including putative mental structures such as scripts, plans, frames, schemas, and mental models (e.g., \cite{Schank&Abelson}, \cite{GentnerMM}), and putative mental processes such as analogy, view application, commonsense perception, analogy, and conceptual combination or conceptual blending (e.g., \cite{WPConBlend}). In AIs driven by Artificial Neural Networks (ANNs), the structures and processes involved in interpretation are diffuse, but these systems engage none-the-less in interpretation.  

Interpretation is also central to Colby's work on paranoia. Indeed, Colby describes one of his earliest implementations of paranoia as precisely interpretation gone awry: ``A parser takes a linear sequence of words as input and produces a treelike structure [...] The final result is a pointer to one of the meaning structures which the interpretation-action module uses in simulating paranoid thinking for both the paranoid and the nonparanoid modes.''\cite[520]{KC81BBS}

Weizenbaum, who it will be recalled had worked with Colby, explicitly recognized that ELIZA had no interpretive machinery - no way to assign meaning to the content of the conversation. Indeed, he chose the Rogerian framework precisely for the reason that that framework (or at least Weizenbaum's gloss of it) puts almost all of the content work on the patient/user, who presumably has intact interpretive machinery: ``This mode of conversation was chosen because the psychiatric interview is one of the few examples of categorized dyadic natural language communication in which one of the participating pair is free to assume the pose of knowing almost nothing of the real world.''\cite[42]{JW66ELIZA} Weizenbaum even points to this as the thing that needs most to be improved in ELIZA in order to create a fuller conversant: ``In the long run, ELIZA should be able to build up a belief structure [...] of the subject and on that: basis detect the subject's rationalizations, contradictions, etc.''\cite[43]{JW66ELIZA}

Yet, at the same time, interpretation is central to Weizenbaum's work, but not the interpretations that ELIZA makes, which are, per Weizenbaum himself, non-existent, but \textit{the interpretations made by the human interlocutors with ELIZA} (or, as described above, with the gomoku player). 

\section{The Threads Come Together: Interpretation, Language, Lists, Graphs, and Recursion}

Interpretation is the process through which a cognitive agent assigns meaning to experience. In the case of ELIZA (or PARRY) ``experience'' is merely textual interaction with the user. Weizenbaum recognized that ELIZA has no interpretive structures or processes at all. It is, however, worth making a connection between interpretation, and the focus described above on recursion and lists, and language, because this connects all of the threads of research that are described here. To create a completely tied-together coherent explication of this complex knot of concepts would require much more space than is available at the moment, so we will merely touch the topic at a few points which will hopefully be sufficient to give the reader the sense of the whole.

First, observe that at the simplest level, a sentence is a ``merely'' list of words. However, as pointed out by Chomsky, human language is fundamentally recursive, as can easily be seen in either the understanding or production of novel, potentially infinitely deeply nested sentences that may be hard to understand, but which, like this one, are nonetheless grammatical and sensible, for example: ``The rat that the cat that the dog chased killed ate the malt.''\cite[286]{ChomskyMiller1963}

Second, observe that a list (perhaps representing a sentence) is a special case of a graph, and, moreover, a nested list (perhaps representing a nested sentence, like the example above) is just a tree, which is a specific kind of graph. Indeed, Chomsky's underlying grammars are graphs as well, where elements link to other elements. And even without going into that depth, it is obvious that language is graph-structured, not merely tree-structured even on the surface, through, for example, pronouns which refer to other constituents of the sentences, creating edges in the graph.

Third, observe that the belief structures that are used in most AI to represent beliefs are generally graphs, wherein symbols, which might represent concepts, are linked to one another by edges, which might represent relationships between concepts.\footnote{In AIs built out of ANNs the concepts and their inter-relationships are not so concisely represented -- usually being diffused across the network, however, the networks themselves are still graphs (indeed, the term ``network'' is just a synonym for ``graph''), and the analytic bases for the construction and analysis of ANNs rest firmly on graph theory.}  

Fourth, observe that graph (or tree or list) traversal is a naturally recursive process, proceeding from vertex to vertex along the edges of the graph, and that many of the core algorithms of classical ``symbolic'' AI rely upon various versions of efficient graph (or tree) traversal.\footnote{The operation of ANNs relies upon matrix multiplication rather than upon graph traversal, however these are closely related, and, indeed, one commonly implements graph traversal via matrix multiplication. Furthermore, advanced users of ANNs are coming to the realization that in order to understand what the ANN is doing, and to guide it ``intelligently'' we are likely to end up relying upon more classical sorts of algorithms, resulting in a hybrid of ANN and symbolic AI.}

Finally (fifth), observe that extended conversations (discourse), is analogous to individual sentences in that at the surface it is linear (list-like), but even going one level down one finds that real discourse contains explicit connections (e.g., ``...before you said...''), and, again analogous to our story of interpretation, as well at both the levels of semantic and pragmatic connections that connect parts of the conversation to one another. 

Given these glosses, it can be seen (roughly) that interpretation is a process -- indeed, a computable recursive function, precisely in Turing's sense -- that transforms one sort of graph structure, the surface structure of sentences, and indeed of whole discussions, to another -- the meaning -- and then back again, into the next turn of the conversation.\footnote{NN-based AIs, especially modern Large Language Models (LLMs) usually operate word-by-word (more precisely, token by token), generating the next ``most likely'' word (token) as result of bubbling the context -- i.e., the whole previous interaction (including the LLM's own outputs) back through the network in what is called a ``recurrent'' pattern. These are more like ELIZA than they are like the AIs built by Colby, Schank, etc. Instead of an author creating a script, the scripts for LLMs a created by transforming enormous amounts of language, usually scraped from the web, into the incredibly complex graphs describing how words relate to other words in their context, but they have no explicit representation of meaning, aside from the nest of interrelationships burned into and buried in the network. As a result, when engaged in discourse, LLMs act striking like ELIZA in that they can briefly maintain the appearance of understanding but once one attempts to carry on the conversation in a new direction, or refer (directly or indirectly) to previous conversational context, they are essentially as lost as ELIZA, and engage in what might be most charitably described as grammatically correct confabulation.} Although ELIZA only engages in the shallowest such recursive operations in transforming input sentences into output sentences directly in accord with its script, Weizenbaum was explicitly aware that ELIZA's users -- and indeed the users of any AI (or for that matter every human everywhere all the time) -- were engaged in interpretation in all is complexity and glory. Putting this more clearly, the users of an AI are interpreting the program as intelligent, and they are led -- or perhaps mis-led -- to this interpretation by virtue of the program putting forward an \textit{appearance} of intelligence, or what Weizenbaum called ``the illusion of understanding'' in his conversation with McCorduck, and, as we shall soon see, in the ELIZA paper itself. 

It is critical to understand that this ``illusion'' does not arise through triggering some sort of abnormal cognitive error. Quite to the contrary, it relies -- as does all magic -- upon the perfectly normal, continuous, and central cognitive process of interpretation that humans are engaged in all the time in everything they do. Without the continuous cognitive process of interpretation, we could not operate, and indeed, we would not be cognitive agents at all. Mistaken interpretation is a common and normal feature of cognition, and is usually easily corrected, if becomes relevant at all.\footnote{Uncorrected mistaken interpretation maybe be said to be the core problem in some of the cognitive impairments studied by Colby and other psychiatrists.}   

Armed with this insight, we come, finally, to ELIZA itself.

\section{Finally ELIZA: A Platform, Not a Chat Bot!}

In short order, after the Datamation article, Weizenbaum publishes a series of papers in the Communications of the ACM (CACM), the premier publication in computing research and development at that time. His SLIP paper appears in 1963\cite{JW63SLIP}, and just three years later, in 1966, having moved to Stanford and then MIT, all of this comes together into the ELIZA paper.\cite{JW66ELIZA} 

Again, let's carefully read what should be obvious, but is commonly overlooked, the title of the paper: ``ELIZA-- A Computer Program For the Study of Natural Language Communication Between Man And Machine.'' This paper begins with a somewhat less negative version of the same plaint as the gomoku paper: 

\begin{quote}
``Introduction. It is said that to explain is to explain away. This maxim is nowhere so well fulfilled as in the area of computer programming, especially in what is called heuristic programming and artificial intelligence. For in those realms machines are made to behave in wondrous ways, often sufficient to dazzle even the most experienced observer. But once a particular program is unmasked, once its inner workings are explained in language sufficiently plain to induce understanding, its magic crumbles away; it stands revealed as a mere collection of procedures, each quite comprehensible. The observer says to himself ``I could have written that''. With that thought he moves the program in question from the shelf marked ``intelligent'', to that reserved for curios, fit to be discussed only with people less enlightened than he.''\cite[36]{JW66ELIZA} 
\end{quote}

Weizenbaum continues in the next short paragraph: ``The object of this paper is to cause just such a re-evaluation of the program about to be ``explained''. Few programs ever needed it more.''\cite[36]{JW66ELIZA} This last sentence reveals something hidden: ``Few programs needed it more.'' This echo of the gomoku paper tells us that it is exceedingly likely that Weizenbaum had already had the experiences that become so famous much later, of people talking intimately to ELIZA, and which became so antithetical to Weizenbaum, leading him to his later positions on AI. 

He only briefly revisits this later in the paper: ``A large part of whatever elegance may be credited to ELIZA lies in the fact that ELIZA maintains the illusion of understanding with so little machinery.'' But he immediately returns to technical matters ``But, there are bounds on the extendability of ELIZA's ``understanding''...'' And once more, the very last paragraph of the paper reads:

\begin{quote}
``The intent of the above remarks [about translating processors] is to further rob ELIZA of the aura of magic to which its application to psychological subject matter has to some extent contributed. Seen in the coldest possible light, ELIZA is a translating processor [...] which has been especially constructed to work well with natural language text.''  
\end{quote}

Interestingly, this is almost identical to Searle's ``Chinese Room'' argument about AI, which appeared significantly later, in 1980.\cite{SCR80} If any program was ever to work in exactly the version described by Searle, it would be ELIZA! Indeed, Searle specifically mentions ELIZA: ``I will consider the work of Roger Schank and his colleagues at Yale [...]. But nothing that follows depends upon the details of Schank's programs. The same arguments would apply to Winograd's SHRDLU [...], Weizenbaum's ELIZA [...], and indeed any Turing machine simulation of human mental phenomena.'' [italics added] 

Let us pause to take note of how this last phrase of Searle's ties together (under his critique, in this case) another set of threads, this time from Hilbert through Turing and up to ELIZA and its descendants. Under the present hypothesis, Searle and other critics of ELIZA as an AI miss the point of ELIZA; ELIZA was not intended as an AI (orchatbot) at all, but as a platform for experiments in human interaction with AIs, and in particular the problem of interpretation; a problem that is now much more important, and equally under-studied as it was 60 years ago, when Joseph Weizenbaum set out to (by this hypothesis) build a platform for research in this area, even going to far as the describe ELIZA as a vehicle for running a version of Turing's imitation game:

\begin{quote}
``With ELIZA as the basic vehicle, experiments may be set up in which the subjects find it credible to believe that the responses which appear on his typewriter are generated by a human sitting at a similar instrument in another room. How must the script be written in order to maintain the credibility of this idea over a long period of time? How can the performance of ELIZA be systematically degraded in order to achieve controlled and predictable thresholds of credibility in the subject? What, in all this, is the role of the initial instruction to the subject? On the other hand, suppose the subject is told he is communicating with a machine. What is he led to believe about the machine as a result of his conversational experience with it? Some subjects-have been very hard to convince that ELIZA (with its present script) is not human. This is a striking form of Turing's test. What experimental design would make it more nearly rigorous and airtight?''\cite[42]{JW66ELIZA}
\end{quote}

Understanding ELIZA as a platform for research into human-machine communication is a very different framing of ELIZA than is commonly attributed to Weizenbaum. On the one hand, in his 1975 book\cite{JW79CPHR} Weizenbaum frames AI as dangerous and/or immoral, and mentions ELIZA very little. His papers circa ELIZA, around the mid 1960s, as we saw above, describe a weaker, although related framing of ELIZA and other AIs of the time as intentional illusions. 

But there is a third way of framing ELIZA that aligns both of these, and finally draws together all of the threads that we have discussed. This third framing might be rendered as: ``AI could be dangerous, and is possibly immoral, if people interpret it the wrong way. Therefore, we need to study how people interpret their interaction with complex computer programs, especially ones that may appear to be intelligent. ELIZA is a platform in support of that project.'' Under this theory ELIZA is not an AI at all, and is only barely a chatbot. Rather, \textit{ELIZA is a platform for research into how human interpretation works, and potentially how it may go awray or be abused.} Indeed, Weizenbaum himself wrote a detailed outline of a paper discussing potential experiments along these lines that was recently uncovered in Weizenbaum's archives at MIT.\cite{JWELIZAFutureDraft}. That paper, signed in Weizenbaum's own hand, describes experiments he envisioned to be carried out with ELIZA, including detailed discussion of potential experiments in discourse and discourse correction. The very first page of that outline, after a brief introduction, dives right into the heart of the questions we have described in the present paper as the ones that Weizenbaum was most interested in, and which ELIZA was designed to explore:
\begin{quote}
    ``Understanding and misunderstanding. A. Our concern with partial understanding and thus understanding and misunderstanding, in diadic communication
derives from both the clinical observation and the experimentally
demonstratable facts that one or both parties in a two-party
communication can be under the impression that they are understood by the other, while, in fact all that is understood is but a fragment of what is said or typed.''
\end{quote}

That outline was apparently intended to become a paper published alongside the ELIZA paper, but, sadly, as far as we know it never made it beyond the draft stage. And, unfortunately, Weizenbaum never explored the potential potential for ELIZA as an experimental platform. Instead, at least at MIT, ELIZA was primarily explored for its educational potential (also mentioned briefly on the above-described outline), a project carried forward for a time by Paul Hayward, Edmund Taylor, and Walter Daniels.\cite{PH66EduLIZA, ET68EduLIZA}

Although Weizenbaum never himself used ELIZA as a platform for research into human-machine interpretation, some researchers did. In particular, a trio of researchers working right across the Charles river from MIT, at The Stanley Cobb Laboratory for Psychiatric Research, of the Massachusetts General Hospital and Harvard Medical School, explored ELIZA as ``a new research tool in psychology.''\cite{MLQ-2-67} They observed that ELIZA ``allows the stabilization of one party in dyadic communication, for ongoing analysis of certain types of communication and for systematic hypothesis testing about such communication''\cite[190]{MLQ-2-67} Although these researcher were not explicit studying the interpretive aspects of human-computer communication that interested Weizenbaum, it came up regardless: ``The plausible conversation which can be generated by these machines often leads to disputes over the alleged intelligence of the computer. Such disagreement may distract attention from the important contributions such a machine may make to the understanding of communication. For each computer program is a simplified model of dyadic communication and may greatly assist in theory construction and testing.''\cite[165]{QML-1-67}. These researchers were using Weizenbaum's MAD-SLIP version of ELIZA,although with a slightly different script (called ``YapYap'' \cite{ElizagenAntGarfinkelNews}) and recently researchers studying the archives of Harold Garfinkel discovered original transcripts of their subjects' conversations with this ELIZA, and the YapYap script, and demonstrated that this script exactly recreated  the original conversations.\cite{ElizagenAntGarfinkelNews} 

Garfinkel and his coworkers interests seem to have been more aligned with what Weizenbaum had in mind: ``Garfinkel was interested in how human–computer interaction was exploiting human social interactional requirements in ways that not only forced participants to do the work of making sense of a chatbot’s turns, but also gave them the feeling of an authentic conversation.''\cite{EMTR23} Unfortunately, Garfinkel's research in this area was never published, and research with ELIZA, and indeed all research into the important interpretive phenomena that interested, and later horrified Weizenbaum, appears to have ended at this point.\footnote{With the exception of the Hayward et al. educational research, mentioned previously.\cite{PH66EduLIZA} This work was carried out on a highly modified ELIZA, and with an explicit educational goal, again not directly addressing the interpretive problem in human-computer communication.}\footnote{Garfinkel and his coworkers did take this topic up quite directly, and worked for a time with the MGH team, although the Garfinkel-related research ended up using a different platform, called LYRIC, for their work. And, again, was never published.\cite{EMTR23}}

\section{A Perfect Irony: A Lisp ELIZA Escapes and is Misinterpreted by the AI Community}

Our thesis is that Joseph Weizenbaum, afraid of people misinterpreting computers and AI, developed ELIZA not as an AI, but as a platform to support research about people's interpretive process, perhaps more specifically their interaction with AIs. Weizenbaum, of course, had to provide an example -- the DOCTOR script -- to demonstrate the platform, but, as above, he had no illusion that it was actually intelligent, nor even very easily interpreted as intelligent; Much more ``intelligent'' programs already existing in various game playing programs (even his own, from the Datamation paper!), in Colby's Parry, and in other programs that solved math word problems\cite{bobrowStudent}, did database lookups from natural language queries\cite{baseball}, and others. Although some of the sort of research envisioned by Weizenbaum was carried out by the MGH team, a specific event thwarted this use case for ELIZA, and, in almost the deepest possible irony, led to exactly the conflation of ELIZA with AI that Weizenbaum set out to study, not to bring into being.  

Weizenbaum built his ELIZA in MAD-SLIP on the IBM 7090, which was the primary machine at MIT's Project MAC, on the 5th-through-9th floors of Tech Square. Almost immediately after ELIZA's publication in 1966, Bernie Cosell created a Lisp knock-off of ELIZA, based on the algorithm and DOCTOR script in Weizenbaum's paper. Cosell worked at Bolt Beranek and Newman (BBN), a large RAND-like consulting company situated nearby MIT, that commonly hired MIT graduates. Coincidentally, at almost the same time, BBN was a founding site of the ARPAnet\footnote{At the time what became the internet, and then the web, was called the ARPAnet, and BBN built its core hardware and developed the programs that ran it.\cite{WPARPANET})}, and Cosell's Lisp ELIZA diffused rapidly through that network and across the soon-to-be Lisp-centered world of academic AI. As a result, Cosell's Lisp ELIZA rapidly became the dominant strain, and Weizenbaum's MAD-SLIP version, inaccessible on the ARPAnet, was almost instantly forgotten.

Through interesting and complex criss-crossing histories, described in detail in \cite{BBNHistory}, Danny Bobrow, a recent MIT AI graduate who headed up BBN's AI program, brought McCarthy, and thus Lisp, to BBN, where Cosell was exposed to it, and built an ELIZA knock-off in Lisp as a side-project.\footnote{Coincidentally, Bobrow was also the author of STUDENT, the program mentioned above that solved math word problems.\cite{bobrowStudent}} Cosell reports: ``When I was working on the PDP-1 time-sharing system [...] I thought I would learn Lisp. That spring, Joe Weizenbaum had written an article for Communications of the ACM on ELIZA. I thought that was way cool. [...] He described how ELIZA works and I said, “I bet I could write something to do that.” And so I started writing a Lisp program on [the] PDP-1 system at BBN.''\cite[540]{PS09CAW} He continues, ``I wrote that program and got it up and working. Playing with it was an all-BBN project. [...] It was written, at first, in the PDP-1 Lisp. But they were building a Lisp on the PDP-6 at that point—or maybe the PDP-10. But it was the Lisp that had spread across the ARPANet. So [ELIZA] went along with it [...].''\cite[541]{PS09CAW} 

Once Cosell's Lisp ELIZA hit the academic world via the rapid spread of the ARPANet, Weizenbaum's MAD-SLIP version was no longer relevant, and the name ``ELIZA'' (and the ``DOCTOR'' script), as well as the concept, was, from that point forward, associated with Cosell's Lisp version, although its origin was still correctly attributed to Weizenbaum via the CACM paper, leading to a 50-year-long, community-wide misapprehension that ELIZA had been written in Lisp. In addition to being promulgated by Cosell's knock-off being the one that was most easily available, via the ARPANet, it was a natural confusion because Lisp was rapidly becoming the go-to language of AI. Once Lisp came onto the scene no one thought much again about SLIP, or, for that matter, IPL-V. 

\section{Another Wave: A BASIC ELIZA turns the PC Generation on to AI}

Another defining event in ELIZA's descendancy occurred almost exactly a decade later, in 1977  Creative Computing, one of the magazines that, along with BYTE, was the ``Hacker News'' and ``GitHub'' of the mid-70s personal computer explosion, published an ELIZA knock-off written in BASIC.\cite{SNCC77} This was coincidentally well-timed as it coincided nearly exactly with the so-called ``1977 trinity'': The year that the Commodore Pet, the Apple II, and the TRS-80 all appeared.\cite{WPHPC} Within a few years, millions of computer hobbyists -- neither academics nor otherwise professionally involved with computers -- had personal computers of all sorts, mostly with BASIC as their primary user-level programming language. Not a small number of those hobbyists were interested enough by the possibility of AI -- perhaps with HAL9000, the murderous computer from 2001\cite{WPHAL}, still in recent memory -- to type in this BASIC ELIZA (which was only a couple pages of code), and experiment with it themselves. Because of its simplicity, and the personal computer explosion, this ELIZA begat hundreds of knock-offs through the decades, in every conceivable programming language, making it perhaps the most copied and knocked-off program in history.\footnote{Indeed, I curate a web site, ELIZAGen.org\cite{Elizagen}, dedicated to the history of ELIZA and ELIZA-like programs. In that capacity I am regularly sent new, or newly-discovered knock-offs of one or another of the ELIZA threads, usually my own BASIC ELIZA. Just last week, as I write this in April of 2024, I received an email pointing out a version of my ELIZA that ended up, through channels unknown, in Apple's HyperCard programmable notecard system for the original Apple Macintosh computers.\cite{WPhypercard}} Just as Cosell's Lisp ELIZA was spread by the ARPANet, this BASIC ELIZA, was spread by the wide-spread availability of personal computers. 

As a result of these coincidences, and an inherent interest in AI (or at least in talking with computers) the version of ELIZA that was known in the academic community was Cosell's Lisp version, and the version known to the public was the BASIC version. But until it was rediscovered in 2021, the original MAD-SLIP ELIZA, was forgotten, and had not been seen by anyone for at least 50 years.\cite{findingELIZA,Elizagen}

\section{Conclusion: A certain danger lurks there}

Weizenbaum's goal to use ELIZA as a platform for the study of the human process of interpretation was thwarted by exactly what he did not want to see; The ``kludge'' [in nearly the original sense of the term] supplanted the reality. Instead of using ELIZA as a tool to study interpretation and interaction with AI, it became a cause celebre in-and-of-itself, the DOCTOR script being the only one ever seen, because it was so good for such a simple program -- exactly the opposite of Weizenbaum's point! Moreover, exactly what Weizenbaum did not want to happen with regard to Lisp vs. SLIP came to pass; instead of people using Fortran to build complex AI programs. Nearly everyone in AI, or involved in symbolic and/or list processing turned to Lisp, and SLIP eventually faded away.

In 1950, Alan Turing wrote: 

\begin{quote}
``I believe that in about fifty years’ time it will be possible to programme computers, with a storage capacity of about 109, to make them play the imitation game so well that an average interrogator will not have more than 70 per cent, chance of making the right identification after five minutes of questioning. The original question, ‘Can machines think!’ I believe to be too meaningless to deserve discussion. Nevertheless I believe that at the end of the century the use of words and general educated opinion will have altered so much that one will be able to speak of machines thinking without expecting to be contradicted. I believe further that no useful purpose is served by concealing these beliefs. The popular view that scientists proceed inexorably from well-established fact to well-established fact, never being influenced by any unproved conjecture, is quite mistaken. Provided it is made clear which are proved facts and which are conjectures, no harm can result. Conjectures are of great importance since they suggest useful lines of research.''\cite[442]{AT50Mind}
\end{quote}

A mere decade and a half later, Joseph Weizenbaum wrote:

\begin{quote}
``With ELIZA as the basic vehicle, experiments may be set up in which the subjects find it credible to believe that the responses which appear on his\footnote{Even for the time, it is striking that Weizenbaum constantly refers to ELIZA's interlocutors as male, given that the only example he provides of a conversation with ELIZA is (putatively) with a woman!} typewriter are generated by a human sitting at a similar instrument in another room.''\cite[42]{JW66ELIZA} 
\end{quote}

Regardless of how close this description may seem to Turing's test, Weizenbaum did not build ELIZA to pass that test. Rather he built it to run experiments, including ones akin to the Turing Test, that could be used to study human interpretive processes (especially in the case of artificial intelligence). He believed that this research was critically important, but not merely for the academic reasons that motivated Turing and most AI researchers: 

\begin{quote}
``[The] whole issue of the credibility (to humans) of machine output demands investigation [...] [Important] decisions increasingly tend to be made in response to computer output. The ultimately responsible human interpreter of 'What the machine says' is, not unlike the correspondent with ELIZA, constantly faced with the need to make credibility judgments. ELIZA shows, if nothing else, how easy it is to create and maintain the illusion of understanding, hence perhaps of judgment deserving of credibility. A certain danger lurks there.''\cite[42-43]{JW66ELIZA}
\end{quote}

The problem of how humans impute agency, correctness, and intelligence to machines is not only still present, but has become exponentially more important in recent years, with the widespread diffusion of internet bots and large language models. Our modern computational lives might have been better had Weizenbaum pursued his goal of using ELIZA to study people's interpretive interaction with computers, and especially with AIs. Unfortunately, his fear that ``[t]here is a danger [...] that the example will run away with what it is supposed to illustrate''\cite[43]{JW66ELIZA} was too prescient.

\pagebreak
 
\section{Acknowledgements}

This paper developed through extensive discussions over several years with ``Team ELIZA'', including Anthony Hay, Art Schwarz, Sarah Ciston, Peggy Weil, Peter Millican, David Berry, and Mark Marino. I am especially indebted to Anthony Hay and Art Schwarz for detailed discussion about technical aspects of ELIZA and SLIP, such as how symbol manipulation worked in SLIP. Terry Winograd and John Markoff participated in helpful background discussions, and Anne Rawls and Clemens Eisenmann provided access to the Garfinkel archives, and Andrei Korbut introduced us to them. Many anonymous commenters in the Hacker News community helped clarify concepts, and provided additional pointers to information that I could not have easily discovered on my own. I am especially indebted to the MIT Archivists, Myles Crowley, Mikki Macdonald, and Allison Schmitt.

\pagebreak
\printbibliography
\end{document}